\documentclass[a4paper]{article}
\usepackage{xcolor}
\usepackage{INTERSPEECH2019}
\usepackage{cite}
\title{Multi-Graph Decoding for Code-Switching ASR}
\name{Emre Y\i lmaz$^1$, Samuel Cohen$^1$, Xianghu Yue$^1$, David van Leeuwen$^2$, Haizhou Li$^1$}
\address{$^{1}$ Dept. of Electrical and Computer Engineering, National University of Singapore, Singapore \\
         $^{2}$ ICIS, Radboud University, Nijmegen, Netherlands}
\email{emre@nus.edu.sg}

\begin{document}

\maketitle
\begin{abstract}
In the FAME! Project, a code-switching (CS) automatic speech recognition (ASR) system for Frisian-Dutch speech is developed that can accurately transcribe the local broadcaster's bilingual archives with CS speech. This archive contains recordings with monolingual Frisian and Dutch speech segments as well as Frisian-Dutch CS speech, hence the recognition performance on monolingual segments is also vital for accurate transcriptions. In this work, we propose a multi-graph decoding and rescoring strategy using bilingual and monolingual graphs together with a unified acoustic model for CS ASR. The proposed decoding scheme gives the freedom to design and employ alternative search spaces for each (monolingual or bilingual) recognition task and enables the effective use of monolingual resources of the high-resourced mixed language in low-resourced CS scenarios. In our scenario, Dutch is the high-resourced and Frisian is the low-resourced language. We therefore use additional monolingual Dutch text resources to improve the Dutch language model (LM) and compare the performance of single- and multi-graph CS ASR systems on Dutch segments using larger Dutch LMs. The ASR results show that the proposed approach outperforms baseline single-graph CS ASR systems, providing better performance on the monolingual Dutch segments without any accuracy loss on monolingual Frisian and code-mixed segments.
\end{abstract}

\section{Introduction}
\label{sec:intro}
Spontaneous use of words from two (or more) languages in a single conversation, also known as code-mixing or code-switching (CS), is mostly noticeable in multilingual societies~\cite{weinreich1953,myers1989,auer1998,muysken2000,thomason2001,bullock2009}. This type of contact-induced language change is common in: (1) minority languages influenced by the majority language or (2) majority languages influenced by \textit{lingua francas} such as English and French. 

West Frisian (Frisian henceforth) has approximately half a million bilingual speakers and these speakers often code-switch between the Frisian and Dutch languages in daily conversations. The FAME! project\footnote{www.fame.frl} provides a platform to study the impact of such spontaneous language alternation on modern automatic speech recognition (ASR) systems with the goal of building a robust ASR system that can handle this phenomenon.

Impact of CS and other kinds of language switches on the ASR systems has recently received research interest, resulting in multiple approaches for CS acoustic and language modeling~\cite{lyu2006,vu2012,wu2015,weiner2012,lyu2013,li2012,adel2013,yilmaz2016_2,westhuizen2017,singh2018}. The main focus of the FAME! project has been the development of a robust ASR that can recognize CS speech with an emphasis on the low-resourcedness of target Frisian language \cite{yilmaz2018,yilmaz2016_4}. Particularly, this project aims to build a spoken document retrieval system for the disclosure of the archives of Omrop Frysl\^{a}n (Frisian Broadcast) covering a large time span from 1950s to present and a wide variety of topics. The recordings in the local broadcaster's archive contain monolingual Dutch and Frisian speech as well as code-mixed Frisian-Dutch speech.

In our latest work, we have described several acoustic and textual data augmentation techniques for improving the ASR and CS detection performance due to the limited acoustic and text resources of Frisian~\cite{yilmaz2018_1,yilmaz2018_2}. The acoustic data augmentation relies on available monolingual acoustic resources from the high-resourced mixed language (Dutch). Using more monolingual Dutch speech for acoustic model training has found to be effective in improving the general ASR performance, only after increasing the in-domain CS data applying the semi-supervised techniques described in \cite{yilmaz2018,yilmaz2017_2}. 

Similarly, the proposed textual data augmentation aims to increase the amount of available CS text to train a CS bilingual language model (LM) that can more flexibly hypothesize language switches. On account of the large amount of created CS text, the final bilingual LM is trained on a corpus mostly consisting of generated CS text. Despite the overall performance improvements in both monolingual and code-mixed segments, the performance of the CS ASR on the high-resourced Dutch language is still considerably lower than a state-of-the-art monolingual Dutch ASR system. The main reason behind this performance gap is the necessity of using a Dutch LM that is comparable to the Frisian LM in size (either using comparable amount of text or by interpolating with a much smaller weight) in a conventional single-graph setting, not to hamper the recognition of monolingual Frisian and code-mixed utterances which mostly contain Frisian words. 

In this work, we propose a new multi-graph back-end for CS ASR in which a separate search space is employed in the form of parallel graphs for each recognition task to address this shortcoming. Previously, several system combination techniques have been proposed that aim to combine ASR systems at different levels such as language models, lattices or N-best lists for improved recognition performance \cite{fiscus1997,mangu2000,fugen2003,chen2006,lecouteux2012}. The idea of using parallel language-dependent graphs each created using a language-dependent acoustic model has been presented in \cite{hazen2001}. To the best of authors' knowledge, there has been no previous work attempting to incorporate multiple search spaces with a unified acoustic model for CS ASR. In our scenario, we use two monolingual graphs, one for Frisian and one for Dutch speech, and a bilingual CS graph for code-mixed speech. We further describe a designated LM rescoring scheme which uses a graph label that indicates the graph used for hypothesizing the decoding output and performs rescoring by updating the corresponding language model score. We apply the modified CS ASR scheme to the Frisian-Dutch CS corpus to investigate the impact of using competing graphs in a single decoding stage on the ASR accuracy on monolingual segments.  
%\vspace{-0.1cm}
\section{Frisian Language and FAME! CS Speech Corpus}
\label{sec:database}
%\vspace{-0.1cm}
Frisian belongs to the North Sea Germanic language group, which is a subdivision of the West Germanic languages. Historically, Frisian has similarities with Old English while modern Frisian language is under the growing influence of the Dutch language due to long-lasting and intense language contact. Frisian is spoken by half a million speakers. All speakers of Frisian are at least bilingual, since Dutch is the main language used in education in Frysl\^{a}n.

The bilingual FAME! speech database, which has been collected in the context of the \textit{Frisian Audio Mining Enterprise} Project, contains radio broadcasts in Frisian and Dutch. This bilingual data contains Frisian-only and Dutch-only utterances as well as code-mixed utterances with inter-sentential, intra-sentential and intra-word CS \cite{myers1989}. These recordings include language switching cases and speaker diversity, and have a large time span (1966--2015). The content of the recordings is very diverse, including radio programs about culture, history, literature, sports, nature, agriculture, politics, society and languages. For further details, we refer the reader to \cite{yilmaz2016}.

\begin{figure}[t]
  \centering
  \includegraphics[trim=0cm 1cm 0cm 0cm, width=0.45\textwidth]{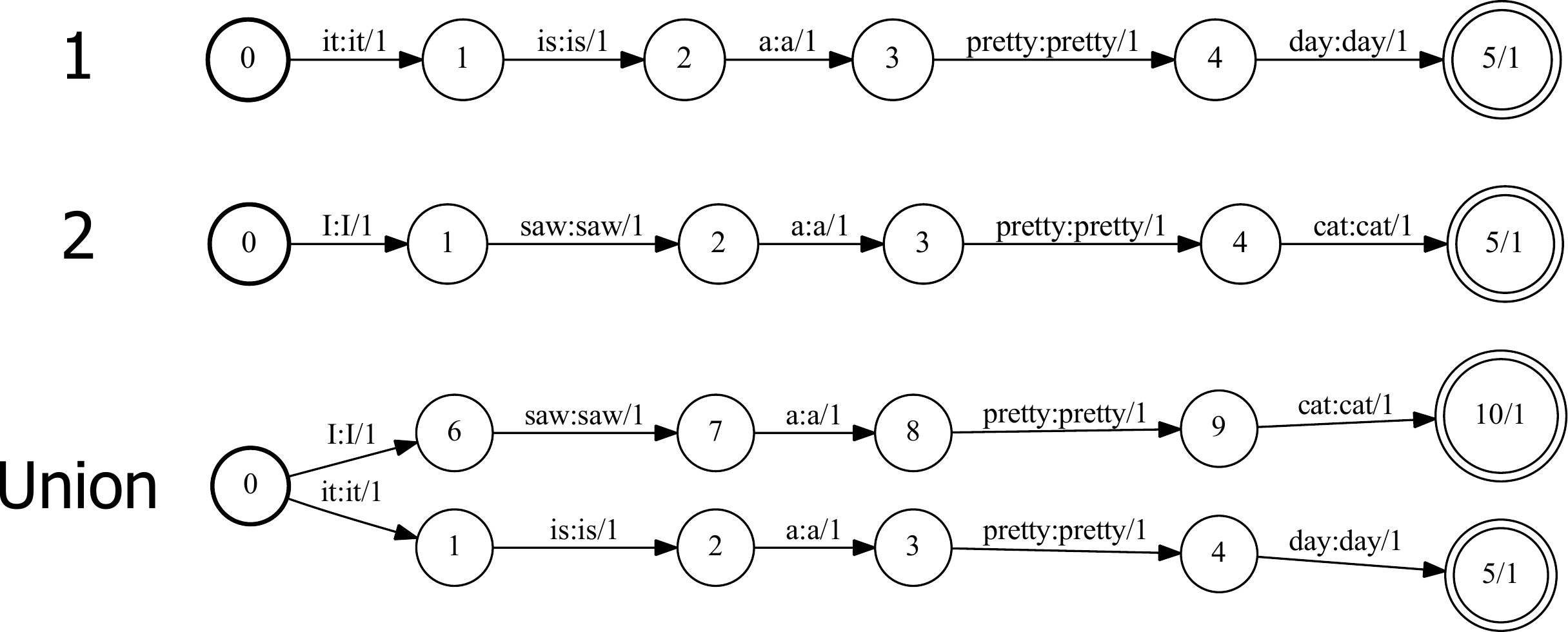}
  \caption{Union of two WFSTs}
  \vspace{-0.5cm}
  \label{fig:union}
\end{figure}
\section{Multi-Graph Decoding}
\label{sec:prop}
%\vspace{-0.1cm}
Modern ASR systems make use of weighted finite state transducers (WFST) \cite{mohri2002,povey2012,yamamoto2012} as a unified representation of different ASR resources with the aim of constructing a mapping from hidden Markov model (HMM) distributions to word sequences. The main motivation is the existence of efficient search algorithms operating on WFST that speed up the decoding process. 

The search space is represented as $(H\circ C\circ L\circ G)$ in the Kaldi toolkit~\cite{kaldi} which consists of the composition of four main components: (1) a grammar WFST (G) with the word sequence information extracted from the LM, (2) a lexicon WFST (L) with the mapping between the orthographic and phonetic transcriptions stored in the pronunciation lexicon, (3) a WFST modeling the phonetic context-dependencies (C), and (4) an HMM WFST containing the HMM topology. The composition ($\circ$) of each component is followed by determinization and minimization operations for reducing the redundancy in the composed graph~\cite{povey2012}.

For our multi-stage decoding technique, we use the union operation for creating a larger graph with parallel bilingual and monolingual subgraphs. The union operation, which is demonstrated with two simple WFST in Figure \ref{fig:union}, combines all the trajectories of two transducers into a single transducer.~The most likely hypotheses from these parallel graphs compete in a shared beam search for the final best path. Equal weights for all subgraphs are used during the union operation. Investigating the effect of setting/learning separate weights for each subgraph remains as a future work. 

The parallel graphs used during the decoding are characterized by the incorporated LM component, as all graphs share the same lexicon, context-dependency and HMM components. For each recognition task, a designated graph is created to be included in the final search graph. Particularly, a monolingual Frisian and three monolingual Dutch graphs are compiled in addition to the baseline CS graph. The former graphs are created using a monolingual LM, while the latter constitutes a CS LM with a large amount of generated CS text together with a smaller amount of monolingual text from both languages. Including some monolingual text in the CS LM is crucial as language switches are often preceded and/or followed by monolingual word sequences.

The Dutch LMs are trained on different text corpora varying in size to explore the impact of LM size on the ASR performance on the Frisian, Dutch and code-mixed utterances. The goal of the multi-graph decoding scheme is to make the recognition accuracy on Frisian and code-mixed utterances less dependent on the size of the Dutch LM compared to standard LM interpolation, providing ASR accuracies on Dutch utterances that are comparable to a monolingual Dutch ASR.

In the multi-graph decoding setting, the LM rescoring has to be performed separately for each graph as each graph uses a different language model. For this purpose, we use graph id tags to be able to identify the graph that is used for hypothesizing the ASR output. Later, the rescoring is performed using the corresponding recurrent neural network (RNN) LM trained on the same data with the N-grams used in the decoding stage.

\vspace{-0.2cm}
\section{Experimental Setup}
\label{sec:exps}
%%\vspace{-0.2cm}
\subsection{Databases}
\label{ssec:data}
\subsubsection{Acoustic data}
%\vspace{-0.1cm}
The training data of the FAME! speech corpus comprises 8.5 hours and 3 hours of orthographically transcribed speech from Frisian (fy) and Dutch (nl) speakers respectively. The development and test sets consist of 1 hour of speech from Frisian speakers and 20 minutes of speech from Dutch speakers each. The sampling frequency of all speech data is 16 kHz. The total amount of automatically transcribed speech data extracted from the target broadcast archive is 125.5 hours. We refer to this automatically transcribed data as the `Frisian Broadcast' data. The automatic transcription procedure is detailed in \cite{yilmaz2018}.

Monolingual Dutch speech data comprises the complete Dutch and Flemish subsets of the Spoken Dutch Corpus (CGN)~\cite{cgn} that contains diverse speech material including conversations, interviews, lectures, read speech and broadcast news. This corpus contains 442.5 hours of Dutch and 307.5 hours of Flemish data.

\subsubsection{Text data}

The baseline CS LM is trained using a bilingual text corpus (107M words) consisting of monolingual Frisian text (37M words), monolingual Dutch (9M words) text and generated CS text (61M words). The Frisian text is extracted from monolingual resources such as Frisian novels, news articles, Wikipedia articles. The Dutch text is extracted from the transcriptions of the CGN speech corpus which has been found to be very effective for language model training compared to other text extracted from written sources. 

The transcription of the FAME! training data is the only source of CS text and contains 140k words. Textual data augmentation techniques described in \cite{yilmaz2018_1} have been applied to increase the amount of CS text. For creating the larger Dutch graphs, we use two subsets of the NLCOW text corpus\footnote{http://corporafromtheweb.org/} with 100M (nl+) and 300M (nl++) words together with the monolingual Dutch corpus with 9M words which is used in the baseline CS LM.
\begin{table}[t]
\centering
\caption{Perplexities obtained on the Dutch component of the development and test transcriptions}
\addtolength{\tabcolsep}{-3.5pt}
\vspace{-0.2cm}
\begin{tabular}{| l | c | c | c | c |}
\hline
LM        	     &  Graph   &	Total \# Dutch words & Dev. & Test \\
\hline \hline
Baseline CS LM   &   cs     &   9M  & 188.1 &  196.6 \\
\hline
interp-nl+       &   cs-nl+ & 109M  & 177.3 &  183.3 \\
\hline
interp-nl++      &   cs-nl++& 309M  & 176.3 &  182.1 \\
\hline \hline
Baseline NL LM   &   nl     &   9M  & 149.6 &  150.7 \\
\hline
nl+              &   nl+    & 109M  & 133.8 &  126.8 \\
\hline
nl++             &   nl++   & 309M  & 122.6 &  118.8 \\
\hline
\end{tabular}
\label{tab:perp}
\vspace{-0.5cm}
\end{table}
\vspace{-0.15cm}
\subsection{Implementation Details}
\label{ssec:impdet}

The recognition experiments are performed using the Kaldi ASR toolkit~\cite{kaldi}. We train a conventional context-dependent Gaussian mixture model-hidden Markov model (GMM-HMM) system with 40k Gaussians using 39-dimensional mel-frequency cepstral coefficient (MFCC) features including the deltas and delta-deltas to obtain the alignments. These alignments are used for training a TDNN-LSTM~\cite{peddinti2017} acoustic model (1 standard, 6 time-delay and 3 LSTM layers) using lattice-free maximum mutual information (LF-MMI)~\cite{povey2016} criterion. We use 40-dimensional MFCC as features combined with i-vectors for speaker adaptation~\cite{saon2013}. Further details on the acoustic model and pronunciation lexicon are provided in \cite{yilmaz2018_1}.

The language models are standard bilingual 3-gram with interpolated Kneser-Ney smoothing and an RNN-LM~\cite{mikolov2010} with 400 hidden units used for recognition and lattice rescoring respectively. The RNN-LMs with gated recurrent units (GRU)~\cite{chung2014} and noise contrastive estimation~\cite{chen2015} are trained using the faster RNN-LM training implementation\footnote{https://github.com/yandex/faster-rnnlm}. In total, we train 7 LMs : (1) baseline CS LM (cs) trained on bilingual text (107M), (2) baseline monolingual Frisian LM (fy) trained on 37M Frisian words, (3) baseline monolingual Dutch LM (nl) trained on 9M words, (4) larger monolingual Dutch LM (nl+) trained on 109M words, (5) largest monolingual Dutch LM (nl++) trained on 309M words, (6) nl+ interpolated with cs (interp-nl+) and (7) nl++ interpolated with cs (interp-nl++). For the last two LMs, the interpolation weights yielding the lowest perplexity on the  transcriptions of the development set are used. These models are used in single-graph ASR systems to compare the performance with the corresponding multi-graph ASR system using the same amount of training Dutch text data. The perplexities of the CS and Dutch LMs on the Dutch component of the FAME! corpus are summarized in Table \ref{tab:perp}.

\begin{table}
\centering
\caption{WER (\%) obtained on the monolingual utterances in the development and test set of the FAME! Corpus}
\addtolength{\tabcolsep}{-1.7pt}
\vspace{-0.2cm}
\begin{tabular}{| l | c || c c | c c |}
\hline
 \multicolumn{2}{|c|}{} & \multicolumn{2}{c|}{Dev.} & \multicolumn{2}{c|}{Test} \\
\hline
 \multicolumn{2}{|c|}{} & fy & nl & fy & nl \\
\hline
 \multicolumn{2}{|c|}{\# of Frisian words} & 9190 & 0 & 10\,753 & 0 \\
\hline
 \multicolumn{2}{|c|}{\# of Dutch words} & 0 & 4569 & 0 & 3475 \\
\hline\hline
ASR System  & Graph  & \multicolumn{2}{c|}{} & \multicolumn{2}{c|}{} \\
\hline \hline
Baseline CS ASR & cs & 22.7      & 24.6      & 21.9 & 20.9           \\
\hline \hline
fy           & fy    & \bf{22.8} & -         & \bf{21.9} & -         \\
\hline
nl           & nl    &  -        & 19.5      & -         & 17.1      \\
\hline
nl+          & nl+   &  -        & 18.3      & -         & 15.9      \\
\hline
nl++         & nl++  &  -        & \bf{17.8} & -         & \bf{15.8} \\
\hline
\end{tabular}
\label{tab:wer_ideal}
\vspace{-0.6cm}
\end{table}
OpenFst is an open-source library designed to handle the WFST~\cite{openfst}. It is highly optimized to perform most WFST operations including minimization, determinization, union, composition. To perform a union operation on different graphs, we use the openfst implementation of the union function \textit{fstunion}. This function takes as input two WFSTs and outputs their union.
\begin{table*}
\centering
\caption{WER (\%) obtained on the development and test set of the FAME! Corpus}
%\addtolength{\tabcolsep}{-3.7pt}
\vspace{-0.2cm}
\begin{tabular}{| l | c | c || c c c c | c c c c | c |}
\hline
 \multicolumn{3}{|c|}{} & \multicolumn{4}{c|}{Dev.} & \multicolumn{4}{c|}{Test} & Total\\
\hline
 \multicolumn{3}{|c|}{} & fy & nl & fy-nl & all & fy & nl & fy-nl & all & \\
\hline
 \multicolumn{3}{|c|}{\# of Frisian words} & 9190 & 0 & 2381 & 11\,571 & 10\,753 & 0 & 1798 & 12\,551 & 24\,122\\
\hline
 \multicolumn{3}{|c|}{\# of Dutch words} & 0 & 4569 & 533 & 5102 & 0 & 3475 & 306 & 3781 & 8883\\
\hline\hline
ASR System  & Graph(s) & Rescoring & \multicolumn{4}{c|}{} & \multicolumn{4}{c|}{} & \\
\hline \hline
 \multicolumn{3}{|c||}{\textit{Single-graph systems}} & \multicolumn{4}{|c|}{} & \multicolumn{4}{|c|}{} & \\
\hline \hline
Baseline CS ASR & cs   & No        & 22.7 & 24.6 & 32.0 & 24.8 & 21.9 & 20.9 & 31.4 & 22.8 & 23.8 \\
\hline
Baseline CS ASR & cs   & Yes       & 21.7 & \bf{23.3} & 31.2 & 23.7 & 20.5 & \bf{19.4} & 29.6 & 21.3 & \bf{22.5} \\
\hline
interp-nl+   & cs-nl+  & No        & 22.7 & 24.2 & 31.7 & 24.6 & 21.9 & 20.3 & 30.7 & 22.6 & 23.6 \\
\hline
interp-nl+   & cs-nl+  & Yes       & 21.7 & 23.1 & 31.2 & 23.7 & 20.5 & 18.9 & 29.5 & 21.2 & 22.5  \\
\hline
interp-nl++  & cs-nl++ & No        & 22.7 & 24.1 & 31.8 & 24.6 & 21.9 & 20.1 & 30.7 & 22.6 & 23.6 \\
\hline
interp-nl++  & cs-nl++ & Yes       & 21.6 & \bf{23.1} & 31.3 & 23.7 & 20.5 & \bf{18.8} & 29.4 & 21.2 & \bf{22.5} \\
\hline \hline
 \multicolumn{3}{|c||}{\textit{Multi-graph systems}} & \multicolumn{4}{|c|}{} & \multicolumn{4}{|c|}{} & \\
\hline \hline
union-fy     & cs, fy  & No & 22.7 & 24.7 & 32.1 & 24.8 & 21.9 & 21.0 & 31.6 & 22.9 & 23.9 \\
\hline
union-nl     & cs, nl  & No & 22.7 & 22.4 & 32.1 & 24.2 & 22.2 & 19.3 & 32.1 & 22.8 & 23.5 \\
\hline
union-nl+    & cs, nl+ & No & 22.8 & 22.6 & 32.2 & 24.3 & 22.0 & 18.9 & 31.6 & 22.5 & 23.4 \\
\hline
union-nl++   & cs, nl++& No & 22.8 & 21.0 & 32.5 & 23.9 & 22.0 & 17.4 & 30.7 & 22.1 & 23.0 \\
\hline
union-nl++   & cs, nl++& Yes & 21.8 & \bf{20.4} & 31.9 & 23.1 & 20.5 & \bf{16.3} & 29.3 & 20.7 & \bf{21.9} \\
\hline \hline
union-fy-nl  & cs, fy, nl  & No & 22.8 & 22.6 & 32.2 & 24.3 & 22.0 & 18.9 & 31.6 & 22.5 & 23.4 \\
\hline
union-fy-nl+ & cs, fy, nl+ & No & 22.7 & 22.2 & 32.5 & 24.2 & 21.9 & 17.8 & 31.6 & 22.2 & 23.2 \\
\hline
union-fy-nl++& cs, fy, nl++& No & 22.8 & 21.1 & 32.6 & 23.9 & 22.0 & 17.3 & 31.2 & 22.1 & 23.0 \\
\hline
union-fy-nl++& cs, fy, nl++& Yes & 22.2 & \bf{20.7} & 32.2 & 23.4 & 21.2 & \bf{16.3} & 30.1 & 21.3 & \bf{22.3} \\
\hline
\end{tabular}
\label{tab:wer}
\vspace{-0.5cm}
\end{table*}
\vspace{-0.4cm}
\subsection{ASR Experiments}
\label{ssec:exps}
%\vspace{-0.1cm}
In the first set of experiments, the ASR performance of the baseline single-graph ASR systems using cs, interp-nl+, and interp-nl++ LMs are presented. The LMs of these systems use the same amount of Frisian and CS text data, but a different amount of monolingual Dutch text. Secondly, we only use the monolingual graphs, namely fy and nl++, to investigate the best possible recognition accuracies with a monolingual back-end that can be achieved on the monolingual Frisian and Dutch subsets of the development and test transcriptions of the FAME! corpus.

The ASR performance of the bi-graph ASR systems is investigated in the third set of experiments. This can be seen as an intermediate step towards the final ASR systems with three graphs, i.e., one CS and two monolingual graphs, and aims to reveal the contribution of each additional monolingual graph to the ASR performance. Finally, the ASR accuracies of the multi-graph ASR systems with varying Dutch graphs are presented.

All systems are tested on the development and test data of the FAME! speech corpus and the recognition results are reported separately for Frisian only (fy), Dutch only (nl) and code-mixed (fy-nl) segments. The overall performance (all) is also provided as a performance indicator. The word language tags are removed while evaluating the ASR performance. The recognition performance of the ASR system is quantified using the Word Error Rate (WER).
\vspace{-0.2cm}
\section{Results and Discussion}
\label{sec:res}
%\vspace{-0.1cm}

Before reporting the results obtained with the multi-graph systems, we present the ASR results obtained by using each monolingual graph (fr, nl, nl+ and nl++) on the corresponding monolingual segments in Table~\ref{tab:wer_ideal}.~The ASR system using only the Frisian (fy) graph provides a WER of 22.8\% (21.9\%) on the Frisian segments. These results are similar to the recognition accuracies provided by the baseline systems indicating that the ASR performance of the baseline CS system on monolingual Frisian segments is on a par with the results that a monolingual Frisian ASR can achieve. On the other hand, only using the largest monolingual Dutch graph yields a WER of 17.8\% (15.8\%) on the Dutch segments which reveals the largest performance gain that could be obtained on the Dutch segments using a larger Dutch LM.

The ASR results obtained using the multi-graph systems are given in Table \ref{tab:wer}. The upper panel summarizes the number of Frisian and Dutch words in each component of the development and test sets. The middle panel shows the baseline results provided by the single-graph CS system using the baseline CS LM and its interpolated versions with the larger Dutch LMs (cs-nl+ and cs-nl++). The baseline CS ASR system provides a WER of 24.6\% (20.9\%) on the Dutch subset of the development (test) set respectively. Interpolating the largest Dutch LMs brings marginal improvements reducing the WER to 24.1\% (20.1\%). The best results after LM rescoring are obtained using the RNN-LM trained on the bilingual text (107M words) for all single-graph systems. All three systems provide similar overall performance with a total WER of 22.5\%, while the best performance on the Dutch subset is provided by the interp-nl++ system with a WER of 23.1\% (18.8\%).

The recognition results obtained using the ASR system with multiple graphs are presented in the lower panel. Similar to the previous results, using a monolingual Frisian graph together with the CS graph (union-fy) does not bring any improvement on the recognition accuracy of the monolingual Frisian utterances. In contrast to Frisian, the ASR using the cs and nl graphs in parallel (union-nl) has an improved WER of 22.4\% (19.3\%) on the Dutch utterances compared to 24.6\% (20.9\%) of the baseline system. This indicates that the Dutch component in the CS graph can provide improved recognition of Dutch segments when used in parallel to the CS graph with no or marginal accuracy loss in other recognition tasks. Using larger Dutch graphs together with the CS graph further reduces the WER to 21.0\% (17.4\%). Applying LM rescoring as described in Section \ref{sec:prop} using two RNN-LMs (trained on the cs (107M) and nl++ (309M) text corpora) provides a WER of 20.4\% (16.3\%) on the Dutch segments and a total WER of 21.9\% which is the best result reported on the FAME! corpus.

Finally, the tri-graph systems (union-fy-nl, union-fy-nl+, and union-fy-nl++) with different Dutch graphs of varying sizes provide comparable results with the corresponding bi-graph systems which do not use a monolingual Frisian graph. This result is consistent with the ASR results presented in Table \ref{tab:wer_ideal} clarifying the ineffectiveness of the fy graph. The LM rescoring using three RNN-LMs  (trained on the fy (37M), cs (107M) and nl++ (309M) text corpora) results in a reduced recognition accuracy on the fy segments due to the inferior performance of the fy RNN-LM compared to the cs RNN-LM. 

In summary, the multi-graph decoding and rescoring approach brings considerable improvements to the monolingual Dutch performance due to the effective use of monolingual text data which is only available for Dutch in this low-resourced scenario. Creating an alternative search space for monolingual segments appears to be a better strategy compared to the baseline single-graph systems using an interpolated LM trained on the same text corpora. Adding a monolingual Frisian graph has been found to be ineffective as the model trained on the limited resources is well-represented in the CS graph. The future work includes an investigation of more effective ways of creating parallel search spaces to reduce the gap between the reported multi-graph results and the results provided by the ASR systems with monolingual Dutch graphs.
\vspace{-0.2cm}
\section{Conclusion}
\label{sec:conc}
\vspace{-0.1cm}
In this paper, we propose a multi-graph decoding and rescoring scheme for the recognition of speech with code-switching (CS). An alternative search space is created for different monolingual and bilingual recognition tasks that is performed during CS ASR. The goal is to compute scores, which are well-calibrated with respect to one another, for hypotheses stored in separately trained graphs using a shared acoustic model.~For evaluating the effectiveness, the proposed technique is applied to the Frisian-Dutch CS scenario, in which the target Frisian language is low-resourced with limited acoustic and text resources. A monolingual Frisian and three Dutch graphs with varying language model size are created for the multi-graph decoding and the ASR performance of the proposed technique on the monolingual utterances is compared with the baseline CS ASR system using a single CS graph. The ASR results have shown that using multiple graphs in parallel provides considerable improvement in the monolingual ASR performance on the high-resourced language reducing the WERs from 23.1\% (18.8\%) to 20.4\% (16.3\%) on the development (test) sets of the FAME! corpus compared to the baseline single-graph systems.
\vspace{-0.2cm}
\section{Acknowledgements}
\vspace{-0.1cm}

This research is supported by National Research Foundation through the AI Singapore Programme, the AI Speech Lab: Automatic Speech Recognition for Public Service Project AISG-100E-2018-006 and the NWO Project 314-99-119 (Frisian Audio Mining Enterprise). The authors would like to thank Henk van den Heuvel for making this collaboration possible.
\bibliographystyle{IEEEtran}

\bibliography{refs}

\end{document}